\documentclass[journal]{IEEEtran}

\usepackage{graphicx}
\usepackage{subfigure}
\usepackage{epstopdf}
\usepackage[figuresright]{rotating}
\usepackage[tableposition=top]{caption}
\usepackage{threeparttable}
\usepackage{color}
\DeclareCaptionFont{blue}{\color{blue}}
\usepackage{makecell}
\usepackage{amsmath}
\usepackage{mathrsfs}
\usepackage{amsfonts}
\usepackage[linesnumbered,ruled,vlined]{algorithm2e}
\usepackage{algpseudocode}
\usepackage{amsmath}
\usepackage{url}
\usepackage{booktabs}
\usepackage{bm}
\let\OLDthebibliography\thebibliography

\renewcommand\thebibliography[1]{
  \OLDthebibliography{#1}
  \setlength{\parskip}{1pt}
  \setlength{\itemsep}{0pt plus 0.3ex}
}

\begin{document}

\title{StarVQA: Space-Time Attention for \\
Video Quality Assessment}


\author{Fengchuang Xing, Yuan-Gen Wang, Hanpin Wang, Leida Li, and Guopu Zhu


\thanks{F. Xing, Y.-G. Wang, and H. Wang are with the School of Computer Science and Cyber Engineering, Guangzhou University, Guangzhou 510006, China (e-mail: 11200606@e.gzhu.edu.cn, wangyg@gzhu.edu.cn, wanghp@gzhu.edu.cn).}

\thanks{F. Xing is also with the School of Physics and Information Engineering, Guangdong University of Education, Guangzhou 510303, China.}

\thanks{L. Li is with the School of Artificial Intelligence, Xidian University, Xian 710071, China (e-mail: ldli@xidian.edu.cn).}
\thanks{G. Zhu is with the School of Computer Science and Technology, Harbin Institute of Technology, Harbin 150001, China (e-mail: guopu.zhu@gmail.com).}

}


\maketitle

\begin{abstract}
The attention mechanism is blooming in computer vision nowadays. However, its application to video quality assessment (VQA) has not been reported. Evaluating the quality of in-the-wild videos is challenging due to the unknown of pristine reference and shooting distortion. This paper presents a novel \underline{s}pace-\underline{t}ime \underline{a}ttention network fo\underline{r} the \underline{VQA} problem, named StarVQA. StarVQA builds a Transformer by alternately concatenating the divided space-time attention. To adapt the Transformer architecture for training, StarVQA designs a vectorized regression loss by encoding the mean opinion score (MOS) to the probability vector and embedding a special vectorized label token as the learnable variable. To capture the long-range spatiotemporal dependencies of a video sequence, StarVQA encodes the space-time position information of each patch to the input of the Transformer. Various experiments are conducted on the de-facto in-the-wild video datasets, including LIVE-VQC, KoNViD-1k, LSVQ, and LSVQ-1080p. Experimental results demonstrate the superiority of the proposed StarVQA over the state-of-the-art. Code and model will be available at: https://github.com/DVL/StarVQA.
\end{abstract}


\IEEEpeerreviewmaketitle

\section{Introduction}

\IEEEPARstart{F}{or} the last few years, user-generated content (UGC) has shown an explosive growth on major social platforms, such as TikTok, Facebook, Instagram, YouTube, and Twitter \cite{Omnicore2021,Chen2020}. This causes a serious problem in content storage, streaming, and usage. Primarily, a deluge of low-quality videos captured by some amateur videographers in severe environments floods into the Internet. It is an urgent task for video quality assessment (VQA) tools to screen these videos according to their quality. However, evaluating the perceptual quality of in-the-wild videos is extremely hard because neither pristine reference nor shooting distortion is available \cite{Mitra2021}.

Convolutional neural networks (CNNs) have delivered remarkable performance on a wide range of computer vision tasks. For example, some deep CNN-based VQA models were proposed \cite{kim2018deep, Zhang2019,you2019deep,VSFA, MDVSFA,chen2020rirnet,LSVQ,tu2021rapique, wu2021}, yielding promising results on the synthetically distorted video datasets. In \cite{kim2018deep}, DeepVQA employed the deep CNN and aggregation network to learn spatiotemporal visual sensitivity maps. In \cite{Zhang2019}, Zhang et al.
 exploited transfer learning to develop a general-purpose VQA framework. You and Korhonen \cite{you2019deep} used 3D convolution network to extract local spatiotemporal features from small clips in the video. VSFA \cite{VSFA, MDVSFA} utilized the pre-trained ResNet-50 to extract spatial features of each frame. RIRNet \cite{chen2020rirnet} designed a motion perception network to fuse the motion information from different temporal frequencies. PVQ \cite{LSVQ} extracted the 2D and 3D features to train a time series regressor and predict both the global and local space-time quality. Tu et al. \cite{tu2021rapique} proposed to combine the spatiotemporal scene statistics and high-level semantic information for video quality prediction. Such semantic information was also oriented to design the VQA method in \cite{wu2021}. Unfortunately, the above methods still struggle on the performance improvement on the in-the-wild videos \cite{LSVQ, LIVE-VQC, hosu2017konstanz} since both the reference videos and distortion types are not available, and the receptive field of convolutional kernels is limited \cite{arnab2021vivit}.

\begin{figure*}[htbp]
\centering
\includegraphics[scale=0.7]{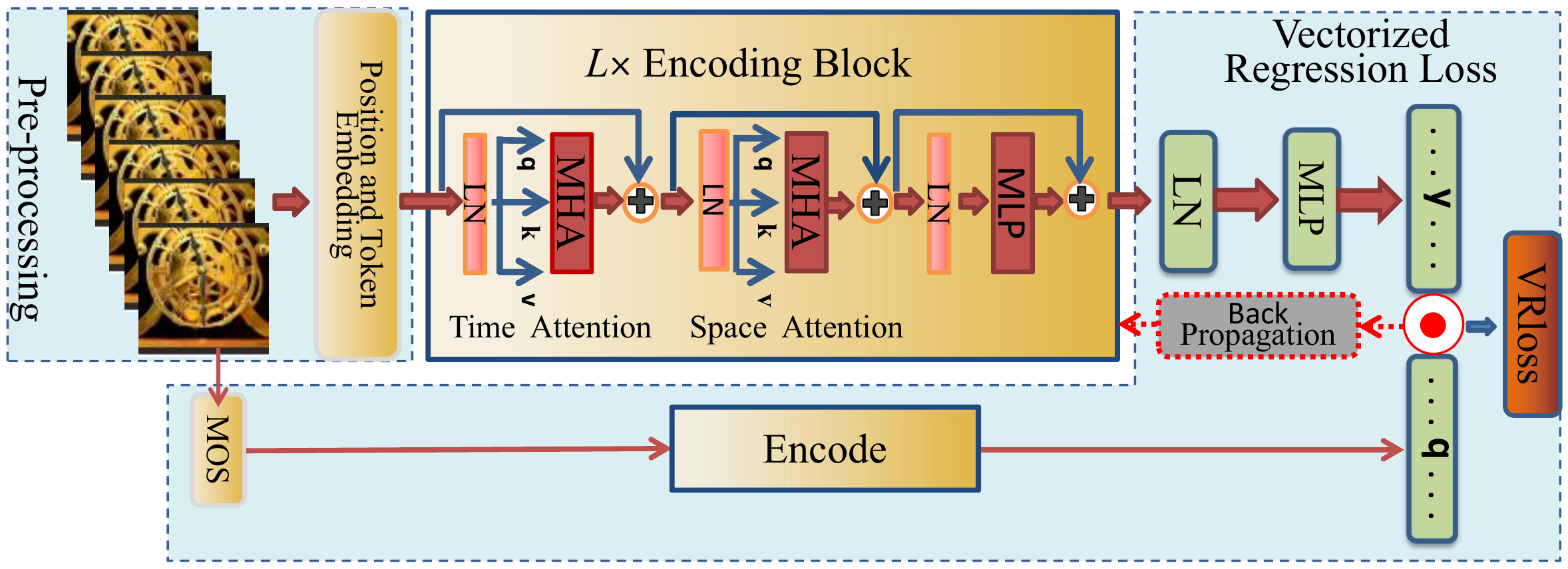}
\caption{Description of StarVQA network. It includes pre-processing, encoding block, and vectorized regression loss modules.}\label{fig1}
\vspace{-0.5cm}
\end{figure*}

The success of attention mechanism in natural language processing (NLP) has recently inspired approaches in computer vision by integrating Transformers into CNN \cite{liu2018end} \cite{wang2018non} or taking the place of CNN completely \cite{ramachandran2019stand} \cite{parmar2018image}. For example, Vision Transformer (ViT) \cite{dosovitskiy2020image} (a pure Transformer-based architecture) has outperformed its convolutional counterparts in image classification tasks. Transformer does not use any convolutions but is based on multi-headed self-attention \cite{Transformer}. This mechanism is particularly effective in modeling the long-term dependency of sequential language. Videos and sentences are both sequential. Thus, one expects that such self-attention will be effective for video modeling as well \cite{timesformer}. Inspired by ViT, several Transformer-based models \cite{arnab2021vivit, timesformer, MViT, motionformer,liu2021,bulat2021} were developed for video classification tasks. These models lead to higher classification accuracy compared with 3D convolutional networks. Unlike the classification tasks that aim at distinguishing among multiple different discrete values, the regression tasks need to output a continuous real value as close to the ground truth as possible. Researches showed that transferring existing classification networks to regression tasks performed not well \cite{dosovitskiy2020image,timesformer,PQA}. As noted by A. C. Bovik \cite{Bovik2020}, ``Unlike human participation in crowdsourced picture labeling experiments like ImageNet, where each human label might need only 0.5-1.0 seconds to apply, human quality judgments on pictures generally required 10-20x that amount to time for a subject to feel comfortable in making their assessments on a Likert scale \cite{Ghadiyaram2016}.'' In general, a clip of video has a long duration including hundreds of images, and its perceptual quality difference from other videos with different content is extremely subtle. Can the Transformer be applied to VQA? If yes, how to implement it effectively? These questions become the original motivation of this work.

In this paper, we design a \underline{s}pace-\underline{t}ime \underline{a}ttention network fo\underline{r} \underline{VQA}, named StarVQA. It is a pure Transformer-based model without using any convolution operations. As a result, StarVQA can capture the long-range spatiotemporal dependencies of a video sequence and guarantee a very fast convergence speed. The major contributions of this work are three-fold: 1) A novel Transformer network is built. To our best knowledge, this is the first work to apply Transformer to the VQA problem. 2) A vectorized regression loss function is designed, which can facilitate the training of StarVQA. 3) Experiments are performed on four benchmark in-the-wild video datasets and show that StarVQA achieves competitive performance compared with five state-of-the-art methods.

\section{Proposed StarVQA}
In this paper, matrices, vectors, and scalar variables are in bold uppercase, bold lowercase, and italic lowercase, respectively. $[N]$, $\lfloor\cdot\rfloor$, and $(\cdot)^T$ denote the integer set $\{1, \dots, N\}$, the floor operation, and  the transpose operation, respectively. The overall framework of the proposed StarVQA is shown in Fig. \ref{fig1}, which consists of pre-processing, $L$ encoding blocks (the divided space-time attention concatenated with a residual connection block), and vectorized regression loss modules.  In the following, we describe each of them in detail.

\textbf{Pre-processing Module.} To match the input of the StarVQA network, we need to pre-process the video sequences. First, we select $F$ frames from each video sequence according to equal-interval sampling, and then crop the selected frame to $H\times W\times 3$ size in a random way, where $H$ and $W$ denote the height and weight of the cropped frame respectively, and the number 3 means the three color channels of R, G, and B. Next, the cropped video frame is divided into many non-overlapping patches with $P\times P$ size. Thus, there are $S=\lfloor H/P \rfloor \times \lfloor W/P\rfloor$ patches in total for each cropped frame. Then each patch is flattened into a column vector with $P\times P\times 3$ dimensions. Denote the column vector
$\mathbf{x}_{(p, t)}\in \mathbb{R}^{P\times P\times 3}$ as the $p$-th patch of the $t$-th selected frame, where $p\in[S]$ and $t\in[F]$.

Since the self-attention mechanism can capture the long-range dependences of spatiotemporal information \cite{timesformer}, we use a spatiotemporal position vector (denoted as $\mathbf{p}_{(p,t)}\in \mathbb{R}^D$ where $D=768$ is set to the dimensions of a patch) to encode each patch $\mathbf{x}_{(p,t)}$ into an initial embedding vector $\mathbf{e}_{(p,t)}^{(0)}$ by
\begin{equation}
 \mathbf{e}_{(p,t)}^{(0)}=\mathbf{M}\mathbf{x}_{(p,t)}+\mathbf{p}_{(p,t)},
\end{equation}
where $\mathbf{M}\in \mathbb{R}^{D\times 3P^{2}}$ denotes a learnable matrix. Next, we add in the first position of the sequence of embedding vectors $\mathbf{e}_{(p,t)}^{(0)}$ for $p\in [S]$ and $t\in [F]$ a special learnable vector $\mathbf{e}_{(0,0)}^{(0)}\in \mathbb{R}^{D}$ representing the embedding of the vectored label token. Finally, we obtain the input of the first encoding block of the StarVQA network (denoted as $\mathbf{E}^{(0)}$) as follows:
\begin{equation}
 \mathbf{E}^{(0)}=\left[\mathbf{e}_{(0,0)}^{(0)}, \{\mathbf{e}_{(p,t)}^{(0)}\}_{p\in [S], t\in [F]}\right].
\end{equation}

\textbf{Time-attention Module.} With the input $ \mathbf{E}^{(0)}$, we can calculate the query  ($\mathbf{q}$), key ($\mathbf{k}$), value ($\mathbf{v}$) vectors of the first encoding block. For each patch, the values of $\mathbf{q}$, $\mathbf{k}$, and $\mathbf{v}$ of the current block $l$ can be successively calculated from the output of the previous block $(l-1)$. For convenience, hereafter $p$ and $t$ can take 0 value  due to an addition of a label token. The calculation process is expressed as
\begin{equation}
 \mathbf{q}_{(p,t)}^{(l,a)}=\mathbf{W}_{Q}^{(l,a)}\textrm{LN}(\mathbf{e}_{(p,t)}^{(l-1)}),
\end{equation}
\begin{equation}
 \mathbf{k}_{(p,t)}^{(l,a)}=\mathbf{W}_{K}^{(l,a)}\textrm{LN}(\mathbf{e}_{(p,t)}^{(l-1)}),
\end{equation}
\begin{equation}
 \mathbf{v}_{(p,t)}^{(l,a)}=\mathbf{W}_{V}^{(l,a)}\textrm{LN}(\mathbf{e}_{(p,t)}^{(l-1)}),
\end{equation}
where $\textrm{LN}(\cdot)$ denotes LayerNorm \cite{layernorm}, $\mathbf{W}_{Q}^{(l,a)}\in \mathbb{R}^{D_h\times D}$, $\mathbf{W}_{K}^{(l,a)}\in \mathbb{R}^{D_h\times D}$, and $\mathbf{W}_{V}^{(l,a)}\in \mathbb{R}^{D_h\times D}$ denote the learnable query, key, and value matrices on the $l$-th encoding block respectively, and $a\in[\mathcal{A}]$ denotes an index over multi-headed attentions and $\mathcal{A}$ denotes the total number of attention heads. The latent dimension for each attention head is set to $D_{h}=D/\mathcal{A}$.

Next, we compute the weight of self-attention. As done in  \cite{timesformer}, we use an alternative, more efficient architecture for spatiotemporal attention, where time-attention and space-attention are separately applied one after the other. First, time-attention is computed by comparing each patch with all the patches at the same spatial location. Thus, we have
\begin{equation}\label{tweight}
 \mathbf{\bm{\alpha}}_{(p,t)}^{(l,a)(\textrm{time})}=\textrm{SM}\left(\frac{\left(\mathbf{q}_{(p,t)}^{(l,a)}\right)^T}{\sqrt{D_h}}\cdot\left[\mathbf{k}_{(0,0)}^{(l,a)}, \left\{\mathbf{k}_{(p,t')}^{(l,a)} \right\}_{t'\in [F]} \right]\right),
\end{equation}
where SM($\cdot$) denotes the softmax activation function. It can be seen from Eq. (\ref{tweight}) that the time-attention coefficient is extracted when $p$ is fixed to a constant. Then, the encoding coefficients can be calculated by using the self-attention weight, which is expressed by
\begin{equation}
 \mathbf{s}_{(p,t)}^{(l,a)(\textrm{time})}={\bm{\alpha}}_{(0,0)(0)}^{(l,a)}\mathbf{v}_{(0,0)}^{(l,a)}+\sum_{t'=1}^{F}\bm{\alpha}_{(p,t)(t')}^{(l,a)}\mathbf{v}_{(p,t')}^{(l,a)}.
\end{equation}
The concatenation of these encoding coefficient vectors from all heads is projected by
\begin{equation}
\begin{split}
 \mathbf{\tilde{e}}_{(p,t)}^{(l)}=\mathbf{W}_{O}^{(\textrm{time})}\left[
 \begin{array}{c}
 \mathbf{s}_{(p,t)}^{(l,1)(\textrm{time})}\\
  \vdots \\
   \mathbf{s}_{(p,t)}^{(l,\mathcal{A})(\textrm{time})}
   \end{array}
   \right]
   +\mathbf{e}_{(p,t)}^{(l-1)},
\end{split}
\end{equation}
where $\mathbf{W}_{O}^{(\textrm{time})}$ is a learnable mapping matrix with $D\times D$ size.

\textbf{Space-attention Module}. To compute the $\mathbf{q}$, $\mathbf{k}$, and $\mathbf{v}$ values of spatial self-attention, we only need to take $\mathbf{\tilde{e}}_{(p,t)}^{(l-1)}$ as the input of $\textrm{LN}(\cdot)$ function of Eqs. (3)-(5). Similar to the temporal self-attention mechanism, the weight of spatial self-attention is computed by
\begin{equation}\label{sweight}
 \mathbf{\bm{\alpha}}_{(p,t)}^{(l,a)(\textrm{space})}=\textrm{SM}\left(\frac{\left(\mathbf{q}_{(p,t)}^{(l,a)}\right)^T}{\sqrt{D_h}}\cdot\left[\mathbf{k}_{(0,0)}^{(l,a)}, \left\{\mathbf{k}_{(p',t)}^{(l,a)} \right\}_{p'\in [S]} \right]\right),
\end{equation}
Therefore, we obtain the encoding weight of spatial self-attention as follows
\begin{equation}
 \mathbf{s}_{(p,t)}^{(l,a)(\textrm{space})}={\bm{\alpha}}_{(0,0)(0)}^{(l,a)}\mathbf{v}_{(0,0)}^{(l,a)}+\sum_{p'=1}^{S}\bm{\alpha}_{(p,t)(p')}^{(l,a)}\mathbf{v}_{(p',t)}^{(l,a)}.
\end{equation}
Like Eq. (8), the output of space-attention can be written by
 \begin{equation}
\begin{split}
 \mathbf{\hat{e}}_{(p,t)}^{(l)}=\mathbf{W}_{O}^{(\textrm{space})}\left[
 \begin{array}{c}
 \mathbf{s}_{(p,t)}^{(l,1)(\textrm{space})}\\
  \vdots \\
   \mathbf{s}_{(p,t)}^{(l,\mathcal{A})(\textrm{space})}
   \end{array}
   \right]
   +\mathbf{\tilde{e}}_{(p,t)}^{(l-1)}.
\end{split}
\end{equation}
Finally, this output is passed through a multilayer perceptron (MLP), using residual connections after each operation:
\begin{equation}
 \mathbf{\bar{e}}_{(p,t)}^{(l)}=\textrm{MLP}\left(\textrm{LN}\left(\mathbf{\hat{e}}_{(p,t)}^{(l)}\right)\right)+\mathbf{\hat{e}}_{(p,t)}^{(l-1)}.
\end{equation}
By now, we obtain the output of the encoding block $l$. And this output will be taken as the input of the encoding block $(l+1)$ until $l=L$.

\textbf{Vectorized Regression Loss Module.} Our attempt to construct the end-to-end Transformer-based networks and use existing loss functions (such as $L$-norm, hinge, and cross-entropy losses) to train such networks \cite{dosovitskiy2020image,timesformer} perform poorly in regression tasks. To adapt the Transformer architecture, we propose to embed a unique vectorized label token to the input of encoding blocks as the learnable variable. Accordingly, the mean opinion score (MOS) is encoded in a vector form. Based on this observation, we design a vectorized regression (VR) loss function for the training of StarVQA.

After obtaining the output of overall encoding blocks, the embedded label token can be taken out to compute the loss. In this module, we employ $\textrm{MLP}$ and $\textrm{SM}$ to generate a probability vector, which is expressed as
\begin{equation}
 \mathbf{y}=\textrm{SM}\left(\textrm{MLP}\left(\mathbf{\bar{e}}_{(0,0)}^{(L)}\right)\right).
\end{equation}
Inspired by \cite{PQA}, we encode the MOS of a video sequence to the probability vector. For this purpose, we first scale the MOS to the interval of [0.0, 5.0]. Then the scaled MOS is encoded to a probability vector $\mathbf{q}=[q_0,\dots,q_5]$ by
\begin{equation}
 {q}_{n}=\frac{{e}^{(-|\textrm{MOS}-\mathbf{b}_{(n)}|^2)}}{\sum_{n=0}^{5}{e}^{(-|\textrm{MOS}-\mathbf{b}_{(n)}|^2)}},
\end{equation}
where $\mathbf{b}=[0,1,2,3,4,5]$ denotes an anchor vector. In Eq. (14), $q_n$ represents the probability of the $n$-th anchor $\mathbf{b}_{(n)}$ corresponding to the MOS. Finally, our VR loss is written as
\begin{equation}\label{loss}
\mathcal{L}_{\textrm{VR}}=1-\frac{\langle\mathbf{q}\cdot\mathbf{y}\rangle}{||\mathbf{q}||\cdot||\mathbf{y}||},
\end{equation}
where $\langle\cdot\rangle$ and $||\cdot||$ denote the inner product operation and $L_2$-norm, respectively. With this loss function, we can train the StarVQA network until convergence. In the last, a linear SVR model can be used to decode the output vector $\mathbf{y}$ and predict the scores of video sequences.


\section{Experimental Results}

\begin{figure*}
\begin{center}
 \subfigcapskip=2pt
    \subfigure[LIVE-VQC]{
    \begin{minipage}{5.5cm}
    \includegraphics[scale=0.35]{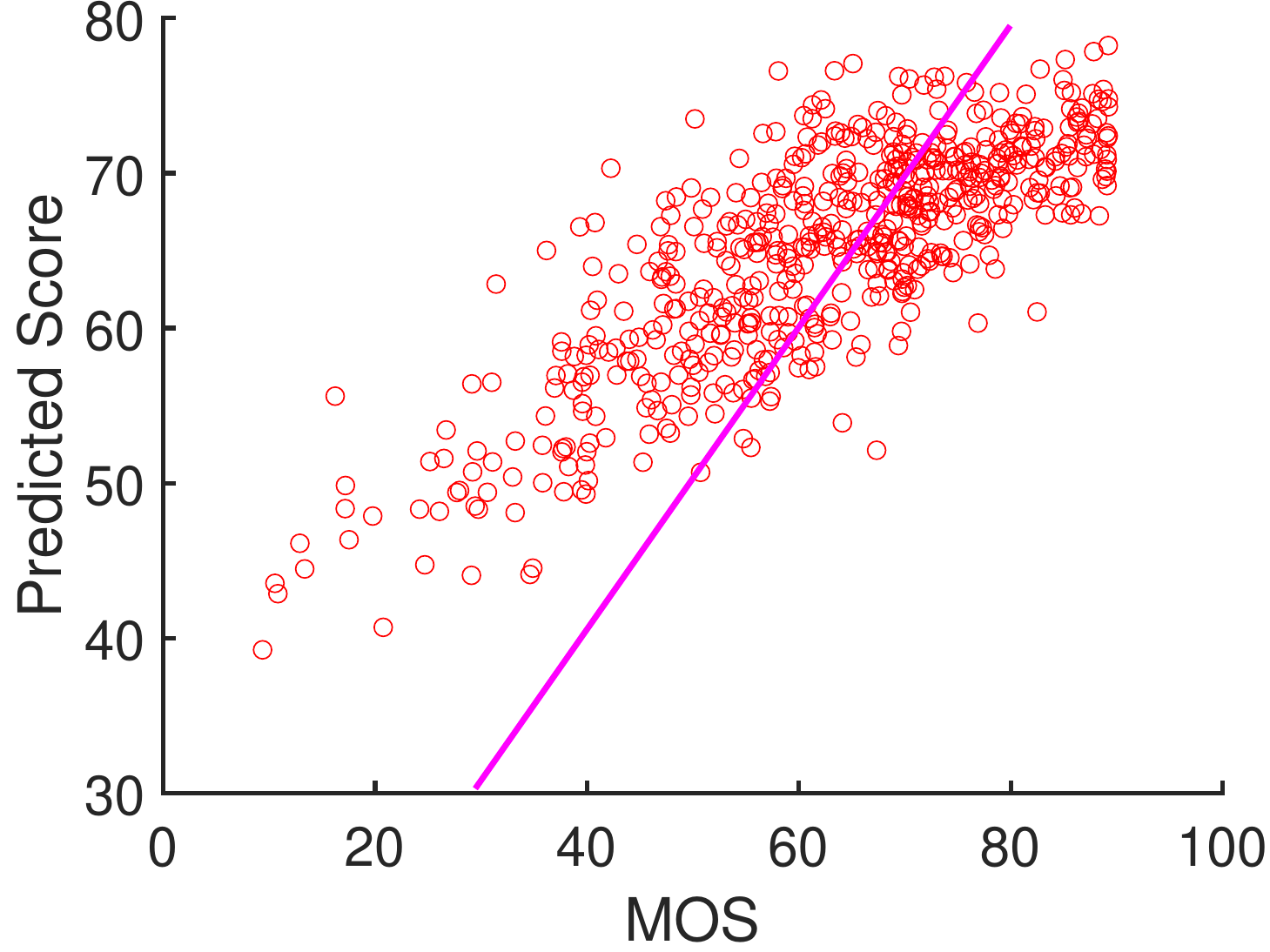}
    \end{minipage}
    }
    \subfigure[KoNViD-1k]{
    \begin{minipage}{5.5cm}
    \includegraphics[scale=0.35]{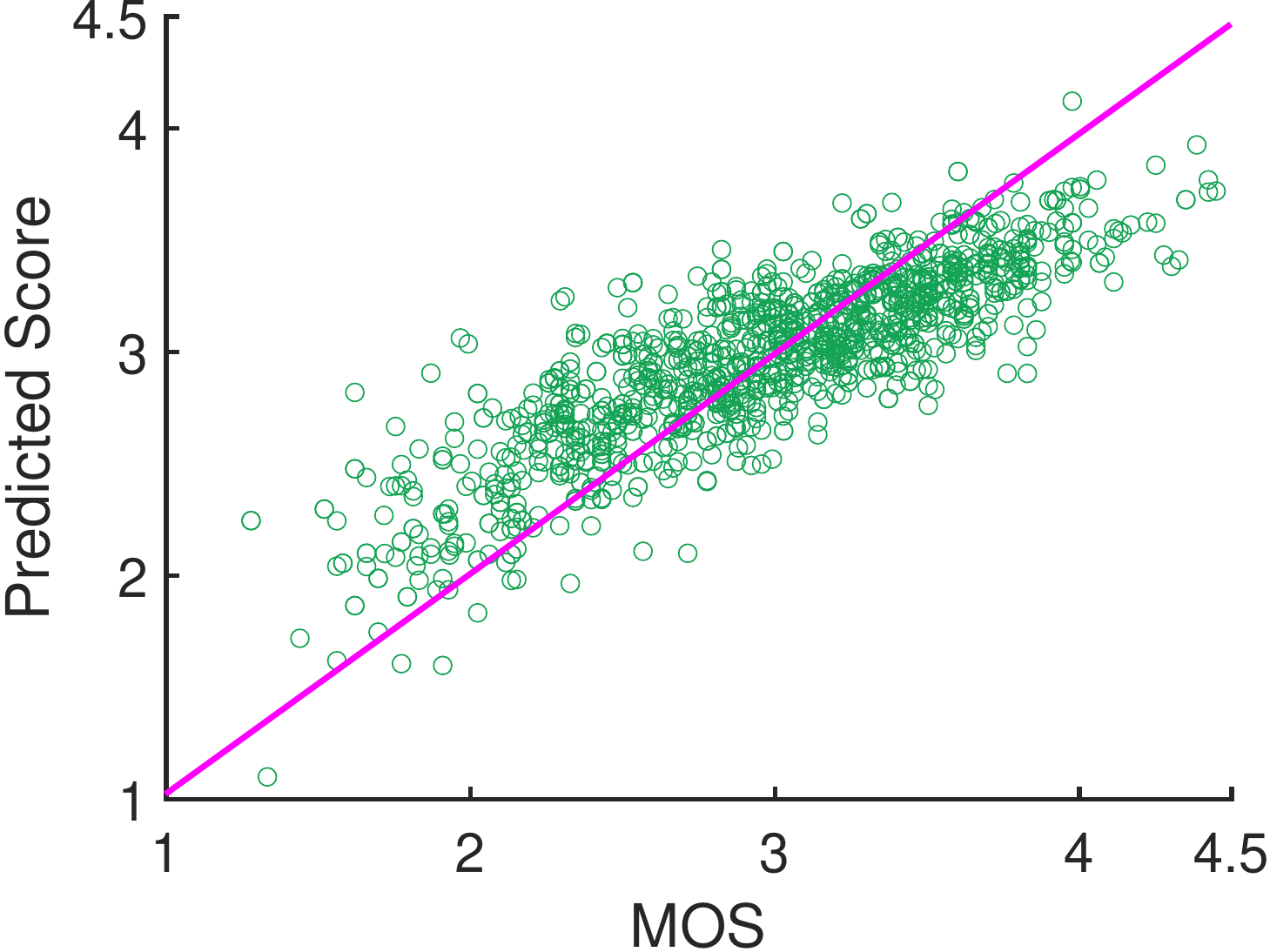}
    \end{minipage}
    }
    \subfigure[LSVQ]{
    \begin{minipage}{5.5cm}
    \includegraphics[scale=0.35]{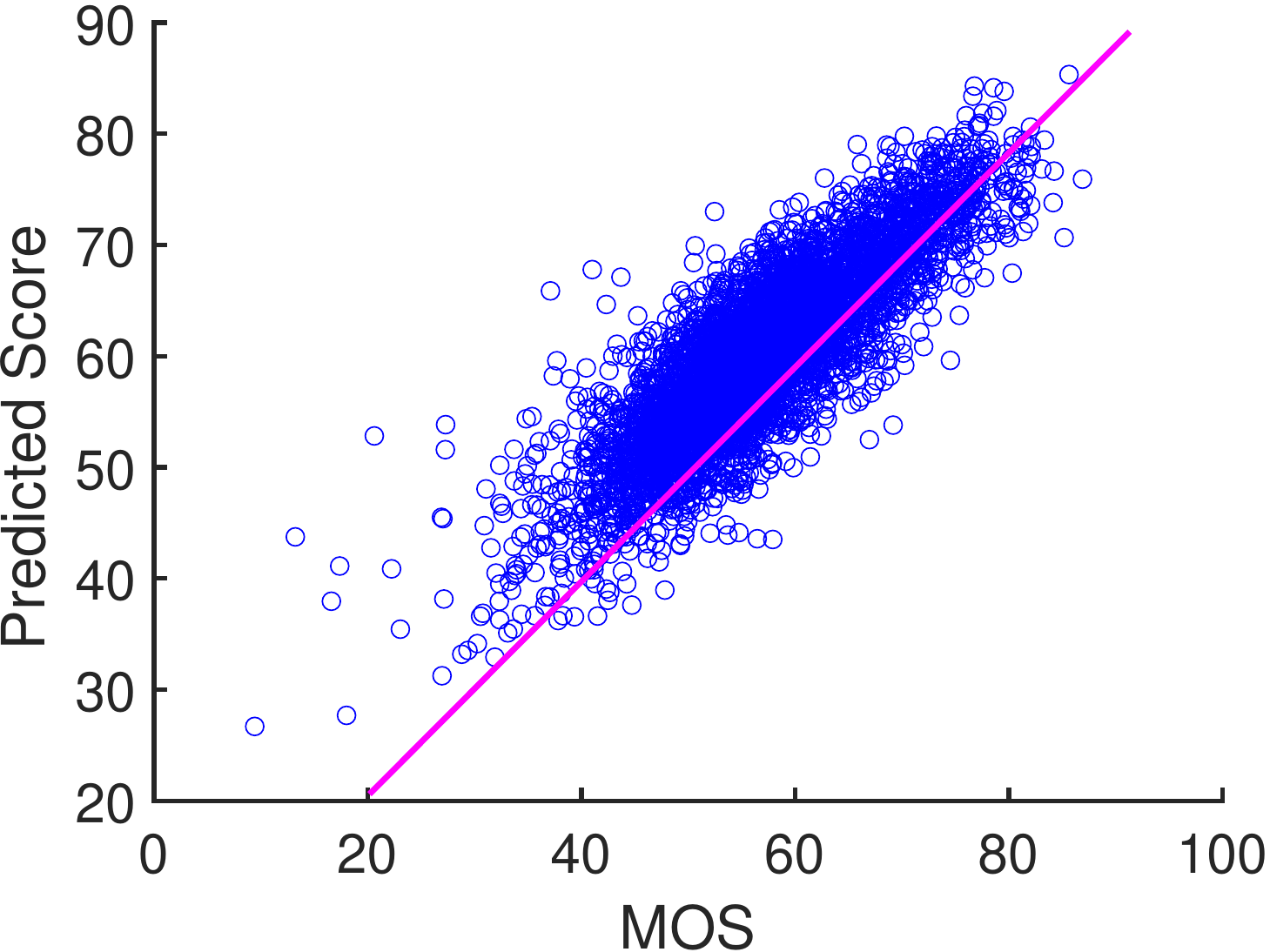}
    \end{minipage}
    }
    \vspace{-0.8em}
    \caption{Scatter plot of predicted scores of StarVQA on different datasets. The magenta solid line represents reference line.}
    \vspace{-0.2em}
    \label{scatter}
    \vspace{-2em}
    \end{center}
\end{figure*}

\textbf{Experimental Setup.} Our network is built on the Pytorch framework and trained on a machine equipped with four Tesla P100 GPUs. Four in-the-wild VQA datasets are used for the verification, including KoNViD-1k \cite{hosu2017konstanz}, LIVE-VQC \cite{LIVE-VQC}, LSVQ \cite{LSVQ}, and LSVQ-1080p \cite{LSVQ}. By convention, $80\%$ of the dataset is for training, and the remaining $20\%$ is for the test. The performance results are averaged on 10 random rounds. Besides, the related parameters are set to $F=8$, $H=W=224$, $P=16$, $\mathcal{A}=12$, and $L=12$.

\begin{figure}[ht]
\centering
\includegraphics[scale=0.5]{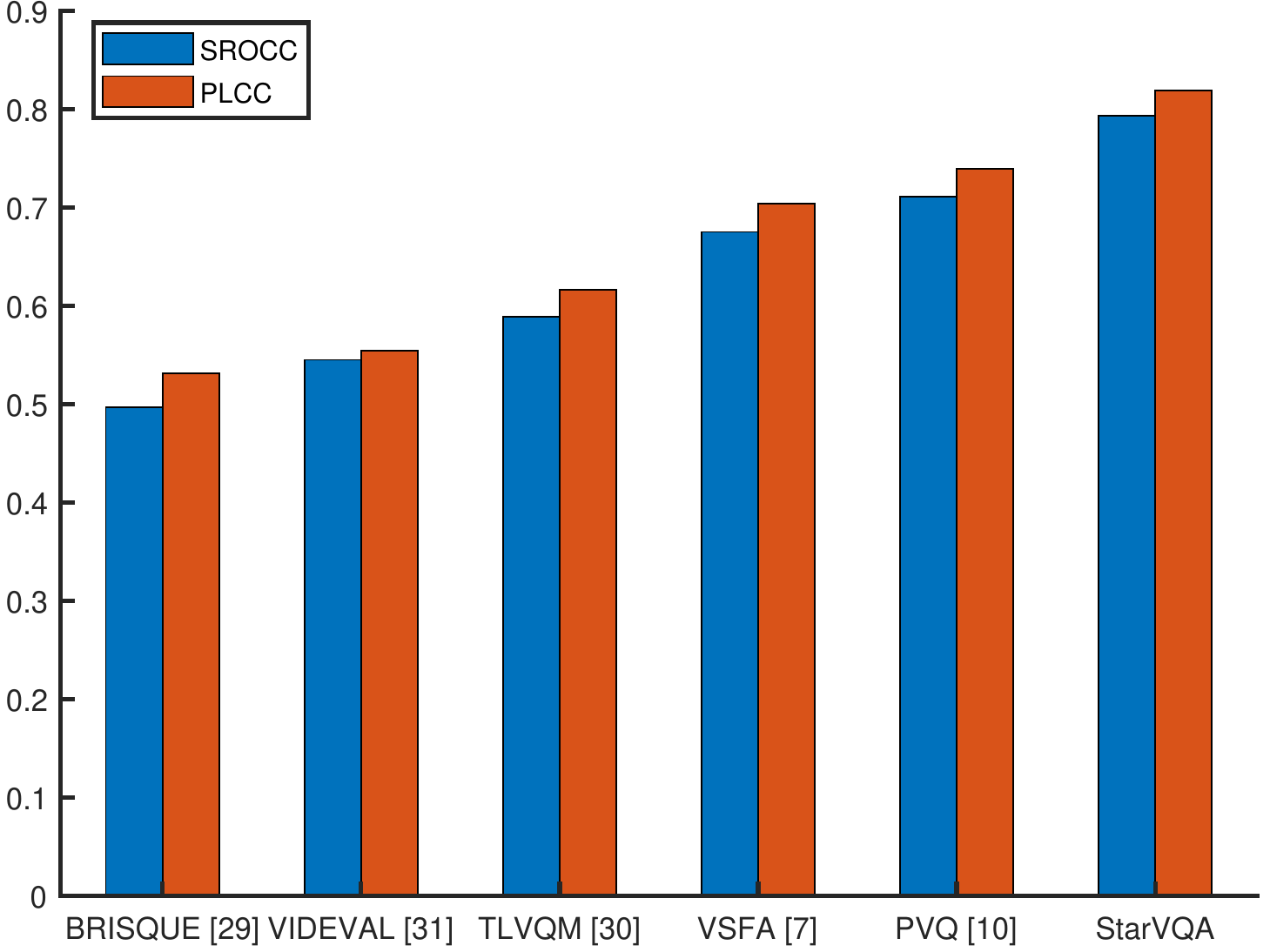}
\vspace{-0.5em}
\caption{Performance verification on high-resolution videos. Here, all the compared methods are pre-trained on LSVQ and tested on LSVQ-1080p.}\label{bar1080p}
\end{figure}

\textbf{Datasets Description.} LIVE-VQC contains 585 videos labeled by MOS of [0.0, 100.0] with resolution from 240p to 1080p. KoNViD-1k contains 1,200 videos labeled by MOS of [0.0, 5.0] with fixed resolution 960p. LSVQ (excluding LSVQ-1080p) contains 38,811 videos labeled by MOS of [0.0, 100.0] with diverse resolutions. LSVQ-1080p consists of 3,573 videos (more than 93\% with resolution 1080p or higher), which are all extracted from the original LSVQ. Note that LSVQ-1080p does not contain any overlapping samples with LSVQ and is specifically designed for the performance verification on high-resolution videos.

\textbf{Convergence Speed.} According to our experiment on the KoNViD-1k dataset, StarVQA achieves 0.72 on both SROCC and PLCC performances when the number of epochs is only five. After the number of epochs exceeds ten, the performance of StarVQA remains almost unchanged. This shows that StarVQA can be well trained at breakneck speed.

\textbf{Overall Performance on Individual Dataset.} The scatter plot of predicted quality scores on different datasets is shown in Fig. \ref{scatter}. We can see that the outputs of StarVQA are all close to the ground truths. This visually demonstrates that the performance of StarVQA remains stable on video sequences from different datasets. Especially, StarVQA on LSVQ gets the most closely centered on the reference line among the three datasets. This is because LSVQ contains a large number of samples, which is most suitable for the Transformer-based networks.

\textbf{Performance Comparison with state-of-the-art.}
In this part, we compare the proposed StarVQA with five state-of-the-art methods, including BRISQUE \cite{BRISQUE}, TLVQM \cite{TLVQM}, VIDEVAL \cite{VIDEVAL}, VSFA \cite{VSFA}, and PVQ \cite{LSVQ}. Comparison results are shown in Table \ref{comparison} and Fig. \ref{bar1080p}. It can be seen from Table \ref{comparison} that StarVQA performs the best on the KoNViD-1k and LSVQ datasets. Nevertheless, our model does not get state-of-the-art performance on LIVE-VQC. This implies that the Transformer architecture may not be very suitable for small datasets. Promisingly, for high-resolution videos, the advantage of the Transformer architecture becomes obvious. From Fig. \ref{bar1080p}, it is clearly shown that StarVQA surpasses all the competitors when pre-trained on LSVQ and tested on LSVQ-1080p.

\textbf{Performance on Cross-dataset.} To verify the generalization of StarVQA, we conduct a cross-dataset experiment. The result is shown in Table \ref{cross}. Note that the comparison result for the cross-LSVQ test on LSVQ-1080p has shown in Fig. \ref{bar1080p}. It can be seen from Table \ref{cross} and Fig. \ref{bar1080p} that StarVQA shows good generalization.

\begin{table}[h]
\caption{Performance comparison with state-of-the-art}
\vspace{-0.5em}
\label{comparison}
\scalebox{0.9}{
\begin{tabular}{|c|c|c|c|c|c|c|}
\hline
  &\multicolumn{2}{c|}{LIVE-VQC}&\multicolumn{2}{c|}{KoNViD-1k}&\multicolumn{2}{c|}{LSVQ}
  \\
  \hline
  Models &SROCC &PLCC &SROCC &PLCC&SROCC &PLCC \\
  \hline
  BRISQUE \cite{BRISQUE} & 0.592&0.638& 0.657&0.658& 0.579&0.576\\
  \hline
  TLVQM \cite{TLVQM} &0.799&0.803&0.773&0.769 &0.772  & 0.774\\
  \hline
  VIDEVAL\cite{VIDEVAL} &0.752&0.751&0.783&0.780&0.794&0.783\\
  \hline
  VSFA \cite{VSFA} &0.773  & 0.795 &0.773  & 0.775&0.801&0.796\\
  \hline
  PVQ \cite{LSVQ} &\bf{0.827}&\bf{0.837}&0.791&0.786 &0.827&0.828\\
  \hline
   StarVQA  &0.732&0.808&\bf{0.812}&\bf{0.796}&\bf{0.851}&\bf{0.857}\\
  \hline
\end{tabular}
}
\end{table}

\begin{table}[h]
\begin{center}
\caption{Performance on cross-dataset}
\vspace{-0.5em}
\label{cross}
\begin{tabular}{|c|c|c|c|c|}
\hline
Training dataset &\multicolumn{4}{c|}{LSVQ}\\
\hline
Testing datasets &\multicolumn{2}{c|}{LIVE-VQC}&\multicolumn{2}{c|}{KoNViD-1k}\\
  \hline
  Models &SROCC &PLCC &SROCC &PLCC \\
  \hline
  BRISQUE \cite{BRISQUE} & 0.524&0.536& 0.646&0.647\\
  \hline
  TLVQM \cite{TLVQM}&0.670  & 0.691 &0.732  & 0.724 \\
  \hline
  VIDEVAL \cite{VIDEVAL} &0.630&0.640&0.751&0.741 \\
  \hline
  VSFA \cite{VSFA} &0.734&0.772&0.784&0.794\\
  \hline
  PVQ \cite{LSVQ} &\bf{0.770}&0.807&0.791&0.795\\
  \hline
  StarVQA  &0.753&\bf{0.809}&\bf{0.842}&\bf{0.849}\\
  \hline
\end{tabular}
\end{center}
\vspace{-0.5cm}
\end{table}

\section{Conclusion}
 Based on the attention mechanism \cite{Transformer}  and TimeSformer \cite{timesformer}, this paper has developed a novel space-time attention network for video quality assessment, named StarVQA. To the best of our knowledge, we are the first to apply Transformer to the VQA field. Furthermore, a new vectorized regression loss function has designed to adapt the Transformer-based architecture for training. Experimental results show that StarVQA achieves  competitive performance compared with five typical VQA methods. This work broadens Transformer to a new application and demonstrates that the attention has excellent potential in the VQA field.

 It is worth mentioning that the result presented in this paper is encouraging. As the first-of-its-kind efforts in the application of Transformer to VQA, great progress may be made with increasing frames selected from a video sequence. The number of frames used in the experiments only takes 8 due to the computation and memory constraints (i.e. 4$\times$Tesla P100 GPUs). According to the result reported in TimeSformer \cite{timesformer}, the accuracy improvement on video classification tasks almost linearly increases as the number of input frames increases. Therefore, we would like to point out that if using clips of 32 or more frames, StarVQA will be a significant departure from current state-of-the-art convolutional models. Besides, we wonder how many encoding blocks used in StarVQA are in a good balance between performance and computation. Under the same computational cost, the convergence comparison of StarVQA with popular CNNs is necessary to be analyzed rigorously. These are worth our further investigation in future.

\bibliographystyle{plain}
\bibliography{starvqa}

\end{document}